\documentclass[letterpaper, 10 pt, conference]{IEEEtran}  % Comment this line out if you need a4paper

\IEEEoverridecommandlockouts                              % This command is only needed if 
\usepackage{cite}
\usepackage{amsmath,amssymb,amsfonts}
\usepackage{algorithmic}
\usepackage{graphicx}
\usepackage{textcomp}
\usepackage{xcolor}
\usepackage{xurl}
\usepackage{booktabs}
\usepackage{float}
\usepackage{adjustbox}
\usepackage{pythonhighlight}
\usepackage{listings}
\usepackage{fancyhdr}
\usepackage{array}
\usepackage[ruled,vlined]{algorithm2e}
\usepackage{stfloats}
\usepackage[most]{tcolorbox}
\usepackage{multirow}

\title{\LARGE \bf
Skill-SLM: Agent Skill-driven Small Language Models for Reliable Robot Operation
}

\author{Wenhao Wang$^{1}$, Yanyan Li$^{2}$, Jiawei Yuan$^{1}$% <-this % stops a space
\thanks{$^{1}$Department of Computer \& Information Science, University of Massachusetts Dartmouth
        {\tt\small \{wwang5, jyuan\}@umassd.edu}}%
\thanks{$^{2}$Department of Computer Science \& Engineering, California State University San Marcos.
        {\tt\small yali@csusm.edu}}%
\thanks{This work was partially supported by the US National Science Foundation awards 2318710 and 2318711, and UMass Dartmouth Internal Research Seed Funding Program.}% <-this % stops a space
}

\begin{document}
\maketitle \thispagestyle{fancy}

%%%%%%%%%%%%%%%%%%%%%%%%%%%%%%%%%%%%%%%%%%%%%%%%%%%%%%%%%%%%%%%%%%%%%%%%%%%%%%%%
\begin{abstract}

Small language models (SLMs) have been increasingly adopted for onboard robot operation because they enable intelligent decision-making. However, existing approaches are mainly distillation-oriented and rely on enumerating representative task-solution pairs. This makes dataset construction difficult and limits generalization to diverse robot tasks whose possible forms grow rapidly. This paper proposes Skill-SLM, a framework that reformulates SLM-driven robot operation as a task-decomposition and skill-composition problem. Given a natural language task instruction, Skill-SLM decomposes the task into subtasks, selects appropriate skills from the skill library, and orchestrates the selected skills into executable robot operations. First, to support the skill-driven workflow, we propose a novel robot operational skill aware context-free grammar (CFG) to extract the skills required to accomplish tasks and build the skill library accordingly. Then, we configure LLM teachers to induce and synthesize training datasets for the SLMs, enabling SLMs to decompose tasks and orchestrate skills reliably. Additionally, we employ a progressive skill orchestration strategy to improve the reliability of skill implementation and overall robot operation. Experiments on UAV operation tasks indicate that Skill-SLM substantially outperforms distillation-oriented baselines, especially on unseen tasks that require generalization of capabilities. Additional experiments on ground vehicle tasks further demonstrate that Skill-SLM can be applied to different robot platforms\footnote{https://anonymous.4open.science/r/SKILL-SLM}.

\end{abstract}

%%%%%%%%%%%%%%%%%%%%%%%%%%%%%%%%%%%%%%%%%%%%%%%%%%%%%%%%%%%%%%%%%%%%%%%%%%%%%%%%

\section{Introduction} \label{sec:intro}

Large language models (LLMs) have enabled robots to interpret human natural language instructions and generate executable actions for navigation~\cite{L3MVN}, planning~\cite{CodeasPolicy}, and manipulation~\cite{manipulation}. However, deploying LLMs on resource-constrained robot platforms (e.g., UAVs) is often impractical due to their substantial computational and memory requirements. As a result, increasing efforts have been focused on enabling onboard deployable language models (i.e., SLMs) to perform reasoning and decision-making reliably through task-specific training and knowledge distillation from LLMs~\cite{Ro-SLM, PRISM, baumann2025enhancing, Small-nav, Small-VLA}.

Despite recent advances, a key challenge remains in enabling SLMs to generalize reliably to diverse robot tasks and novel instructions. Recent studies have enhanced SLMs' generalization ability by extensively enumerating and synthesizing representative task-solution examples~\cite{Ro-SLM, PRISM}. However, these data-driven approaches rely on the SLMs to memorize training examples, which limits their generalization when tasks require unseen combinations of learned capabilities or tasks were insufficiently represented during training. Covering these unseen cases by continually enumerating new task-solution examples is expensive, and SLMs' limited memorization capacity becomes unreliable when learned knowledge becomes large~\cite{scaling-law}. Moreover, it is difficult to anticipate all possible task variations in advance.

Given the compositional nature of robotic operations, recent studies have investigated agent skills~\cite{anthropic-agent-skills} to enable LLMs to generalize to unseen tasks by decomposing them into subtasks and tackling them by composition of skills~\cite{Skills4Robotics, skill-nav}. However, SLMs' intrinsically smaller model scale hinders their ability to achieve reliable compositional reasoning like LLMs~\cite{LLM-scaling}. Specifically, SLMs can be distracted by irrelevant or ambiguous information in the context~\cite{SLM-sensitivity} and struggle to effectively interpret task-relevant context for a given task instruction, resulting in incorrect task decomposition and incorrect skill usage. In addition, SLMs can overlook skill dependencies and fail to coordinate skills into a coherent execution~\cite{slm-skills}. Earlier skills selected by SLMs can define targets, variables, or robot states that later skills must preserve, and incompatible preconditions can make a plausible operational code fail. These limitations highlight the urgent need for methods capable of bridging the broad generalization of compositional reasoning with reliable SLM-driven robot operation.

\begin{figure}[t]
    \centering
    \includegraphics[width=0.85\linewidth]{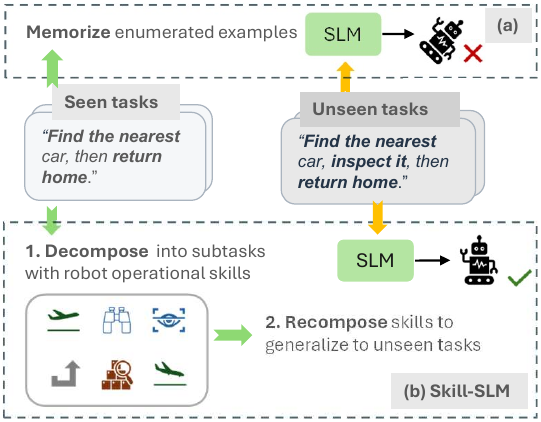}
    \caption{Comparison of SLM-driven robot operation methods: (a) memorizing enumerated examples limits generalization to unseen tasks, whereas (b) Skill-SLM decomposes tasks by operational skills and recomposes them to solve unseen tasks.}
    \label{fig:overall}
\end{figure}

To address these limitations, we propose Skill-SLM, a framework in which SLM agents solve robotic tasks by task decomposition and skill composition, as illustrated in Fig.~\ref{fig:overall}. Skill-SLM consists of two agents: a decomposition SLM agent decomposes a task instruction into subtasks and selects appropriate skills from the skill library; an orchestration SLM agent then orchestrates the selected skills into executable robot operations. Unlike prior methods that map task instructions to complete robot operations end-to-end, the proposed Skill-SLM captures the compositional nature of robotic tasks; it not only improves the seen task performance through compositional reasoning, but also facilitates generalization to diverse robotic tasks.

Specifically, we first construct a skill library to support the skill-driven workflow. We configure an operational skill aware LLM CFG constructor to induce a CFG that captures the skills that can effectively complete the tasks. Based on the captured skills, we construct a skill library with expert knowledge, in which each skill is documented in terms of its functionality, usage, and dependencies. The SLM can refer to these skill documents when orchestrating skills into executable robot operational code. On top of that, we distill LLMs' reasoning to enable the SLM to effectively perform compositional reasoning, we leverage LLMs to synthesize two distillation datasets that capture different stages of the reasoning process. The \emph{decomposition dataset} consists of task instructions, subtask descriptions, and selected skills, which trains the decomposition SLM agent to learn how to decompose tasks into subtasks and select the appropriate skills for each subtask. The \emph{orchestration dataset} contains subtask descriptions, skill specifications, and their corresponding robot operation code, and is used to train the orchestration SLM agent to accomplish each subtask by orchestrating the selected skills into executable robot operation code. Additionally, to improve the reliability of the orchestration agent, we introduce a progressive orchestration strategy that generates executable robot operation code progressively while preserving the task context, intermediate variables, robot states, and cross-skill dependencies. 

We conduct extensive experiments on UAV and ground vehicle operations to evaluate the effectiveness and generalizability of Skill-SLM. Qwen3-8B~\cite{qwen3} achieves \(100\%\) task success rate on Seen-Pattern tasks, \(100\%\) on New-Combination tasks, and \(94\%\) on New-Capability tasks. Skill-SLM also generalizes to ground vehicle operations, achieving \(100\%\), \(95\%\), and \(100\%\) task success rates on the three task sets, respectively. Compared with the baselines~\cite{Ro-SLM, PRISM}, our method consistently achieves higher task success rates, particularly on unseen tasks that require knowledge recombination. These results demonstrate that Skill-SLM is promising for reliable onboard SLM-driven robot operation across different robot platforms.

\section{Related Work} \label{sec:related}

\subsection{Agent Skills}

Several prior studies have explored skill-like representations. SkiROS2~\cite{skiros2} represents robot skills as parameterized procedures with explicit execution conditions. LRLL~\cite{self-learned-skill} proposed to learn skills via exploring and summarizing. These methods show that reusable skills can improve generalization, but they mainly focus on model-learned skills and are restricted to a constrained environment. The recent agent-skills paradigm~\cite{anthropic-agent-skills} provides a modular, file-based knowledge package that can be discovered and loaded by an agent when relevant~\cite{agent-skills}. Therefore, the agent can dynamically acquire and utilize skills as needed in solving different task scenarios, especially for tasks that are unseen in prior training or require knowledge recombination. Inspired by this principle, this studies investigated how a given task can be reformulated as an agent skill-driven problem.

\subsection{Skill Discovery and Characterization}
Prior work has investigated identifying skills. Ref.~\cite{huaman2017taxonomy} derives a taxonomy of manipulation benchmark tasks from existing evaluation protocols, while~\cite{Pantano2022Capability} systematically analyzes robot skill frameworks to identify recurring task--skill--primitive structures and common robot capabilities. More recently, the BEHAVIOR Robot Suite~\cite{BEHAVIOR} analyzes the 1,000 household activities in BEHAVIOR-1K to identify a set of core whole-body capabilities required for everyday manipulation. These approaches provide useful abstractions for characterizing task corpora, but largely rely on manually specified taxonomies, classification criteria, or capability categories. NavNuances~\cite{NavNuance} instead introduces a CFG-based decomposition of navigation instructions, enabling fine-grained analysis of model performance over task units. Its CFG is constructed by starting from a manually designed grammar and iteratively refining it with LLM assistance. In contrast, we leverage the semantic knowledge of LLMs to infer robot operations from the task corpus and automatically construct the CFG, eliminating the need for handcrafted production rules.

\begin{figure*}[t]
    \centering{\includegraphics[width=0.9\textwidth]{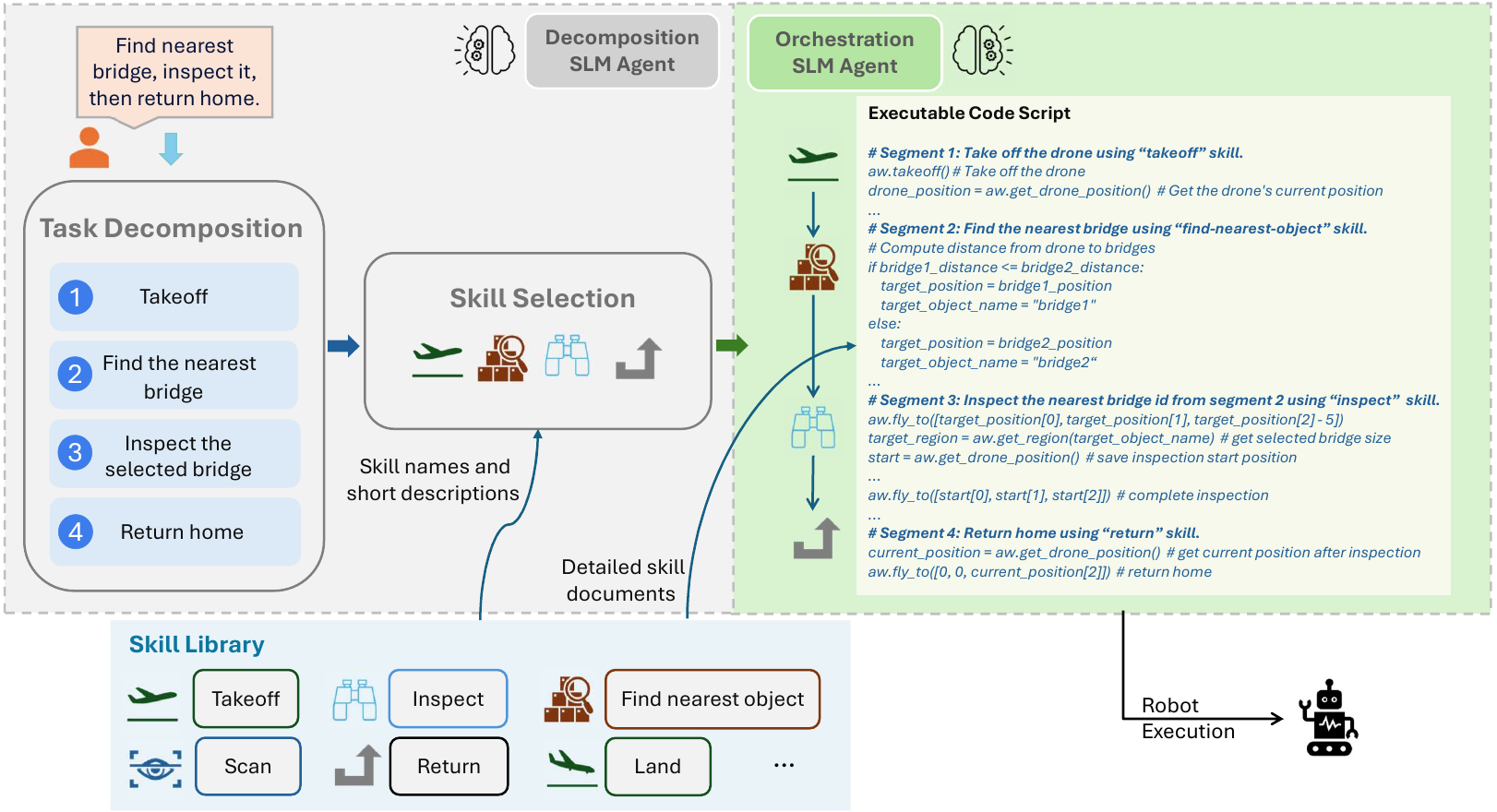}}
    \caption{Overall design of Skill-SLM with an illustrative example.}
    \label{fig:workflow}
\end{figure*}

\subsection{SLM-enabled Robotics}

LLMs have been widely used to connect natural language instructions with robot planning, control, and tool use~\cite{ChatGPTRobotics, CodeasPolicy, gemini-robotics, GSCE}. However, relying on large cloud-hosted models can be impractical for robots operating with unreliable networks (e.g., disaster response) or privacy-sensitive scenarios. This has motivated recent work on SLM-enabled robotics, where smaller models that can be deployed on the robot's computing unit are configured for robotic tasks~\cite{PRISM, baumann2025enhancing, Ro-SLM}. For example, PRISM distills an LLM-enabled planner into an SLM for high-level robot planning~\cite{PRISM} and conducts parameter-efficient training with Low-Rank Adaptation (LoRA)~\cite{lora} to reduce GPU memory usage. Ro-SLM~\cite{Ro-SLM} leverages LLMs to synthesize a dataset and train SLMs for mapping instructions to robot operational code. For a low-level robotic controller, Baumann et al.~\cite{baumann2025enhancing} train SLMs to generate parameters for model predictive control in an autonomous driving system. These approaches demonstrate the feasibility of transferring LLM capabilities to smaller models, but they still largely depend on enumerating representative task-solution mappings for SLMs to memorize. However, such enumeration becomes increasingly difficult as robot tasks become more diverse. Additionally, the small scale of SLMs makes memorization unreliable when the learned facts become large~\cite{scaling-law}. In contrast, Skill-SLM solves tasks by recombining reusable capabilities rather than memorizing all the task solutions.

\section{Method}
\label{sec:method}

\subsection{Overview}

Fig.~\ref{fig:workflow} presents the workflow of our Skill-SLM framework. Given a task \(T\), the goal of Skill-SLM deployed on the robot is to analyze the task and generate the corresponding robot operation code \(C\) for the robot to complete the task. In our framework, the decomposition agent first performs task decomposition and skill selection, and produces a subtask plan:
\[
 p = [(u_1, s_{j_1}), (u_2, s_{j_2}), \ldots, (u_t, s_{j_t})]
\]
where \(u_t\) is the subtask description, \(\mathcal{S}=\{s_1,\ldots,s_K\}\) denotes a library of agent skills, and \(s_{j_t}\in\mathcal{S}\) is the selected skill for the subtask \(u_t\). Then, the orchestration agent performs skill composition by orchestrating \(p\) progressively. At each step \(i\), the agent outputs a code segment \(c_i\), and the final code is obtained by concatenating the segments:
\[
 C = c_1 \oplus c_2 \oplus \cdots \oplus c_t 
\]

\subsection{Skill Library Construction} \label{sec:skill}

\subsubsection{Operational Skill Discovery}

\begin{figure*}[htbp]
    \centering
    \includegraphics[width=0.9\textwidth]{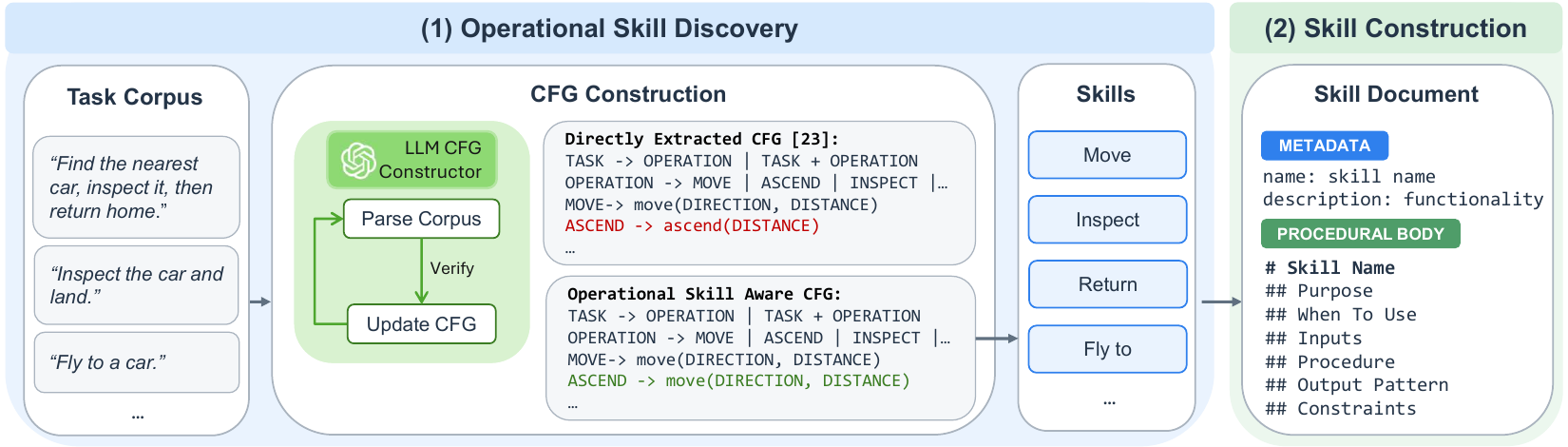}
    \caption{Process of skill library construction.}
    \label{fig:skill-library-construction}
\end{figure*}

To systematically discover required operational skills from task instructions, we leverage the semantic understanding capabilities of LLMs to construct a CFG that captures the structural patterns of task instructions and identifies robot operational skills. However, directly inducing a CFG with an LLM can produce redundant skill definitions by treating semantically equivalent operations as distinct skills. As an example shown in Fig.~\ref{fig:skill-library-construction}, directly extracting a CFG with an LLM produces a grammar that treats the \texttt{ascend} operation as an independent operation, whereas it can be regarded as a specific instance of movement using the \texttt{move} skill. Such redundancy unnecessarily enlarges the skill space and can make skill selection more difficult for SLMs. Therefore, to ensure the completeness and accuracy of the CFG, we configure the LLM to consider both task coverage of the CFG and robot operational semantics.

Fig.~\ref{fig:skill-library-construction} presents the workflow of our design. Given a task corpus, the LLM constructs a CFG that is formalized as a tuple \(G=(\mathcal{N}, \mathcal{N}_s, \Sigma, \mathcal{P}, S_0)\), where \(\mathcal{N}\) is a set of non-terminals representing task instruction components (e.g., \texttt{MOVE}, \texttt{ASCEND}, \texttt{OBJECT}), non-terminals \(\mathcal{N}_s\) representing operational skills (e.g., \texttt{move(DIRECTION, DISTANCE)}), terminals \(\Sigma\) are robot execution elements (e.g., \texttt{turn left}, \texttt{car}), production rules \(\mathcal{P}\) defining how non-terminals can be expanded into sequences of terminals and non-terminals to generate task instructions (e.g., \texttt{MOVE -> move(DIRECTION, DISTANCE)}, \texttt{OBJECT -> car}), and \(S_0\) denotes the start symbol from which the generation of a complete task instruction begins (e.g., \texttt{TASK}). Specifically, the LLM is configured with a system prompt that defines its role as a CFG constructor, specifies the formulation of the CFG, and introduces rules for identifying and consolidating operational skills according to their semantics. The LLM first generates an initial CFG from the task instruction corpus and then performs a verification step to assess whether 1) all task instructions can be represented by the grammar and 2) the identified operational skills are semantically distinct without unnecessary redundancy. Based on this verification, the CFG is iteratively revised if necessary. Finally, the LLM CFG constructor outputs the CFG and the skill list.

\subsubsection{Skill Construction}

Based on the discovered operational skills in \(G\), we construct the skill library. Each skill is grounded in the robot APIs by a human expert and encoded as a structured document:
\[
 d(s)=\bigl(m(s),b(s)\bigr),
\]
where \(m(s)\) is compact metadata and \(b(s)\) is the procedural body. The metadata contains the skill name and a concise description of its functionality. The body provides the SLM with instructions for a skill's implementation; it specifies the skill purpose, conditions of selection, required inputs and preconditions, implementation procedure with relevant robot APIs, expected output pattern, and operational constraints.

\subsection{Progressive Skill Orchestration} \label{sec:orchestration}

For a robot to successfully execute the intended operations, the SLM agent must generate code that is both syntactically correct and logically aligned with the task objectives. However, generating the entire code in a single pass can cause the SLM to forget intermediate operations or use inconsistent variables across skills, making the generated code fail to accomplish the task~\cite{slm-skills}. To mitigate these issues, Skill-SLM introduces a progressive skill orchestration strategy, where the code is orchestrated subtask-by-subtask with respect to the plan \(p\). 

In detail, for the \(i\)-th step in the plan \(p\), the SLM takes \(x_i =(T,u_i,b(s_{j_i}),\mathcal{A},C_{1:i-1})\) as input and generates the code segment \(c_i\) for the current subtask. Here, the task instruction \(T\) preserves the global task objective, while the subtask description \(u_i\) specifies the local objective of the current step. \(T\) and \(u_i\) allow the SLM to maintain the global task intent while focusing on the local requirements of the current subtask. The procedural body \(b(s_{j_i})\) of the selected skill provides the procedural knowledge for implementing the subtask, and \(\mathcal{A}\) specifies the definitions of the robot's available actions to instantiate the skill procedure as executable code. In addition, \(C_{1:i-1}=c_1\oplus\cdots\oplus c_{i-1}\) contains the code generated for all preceding subtasks. Conditioning the generation of \(c_i\) on \(C_{1:i-1}\) enables the SLM to reference previously defined variables, targets, and intermediate results when implementing the current skill. Altogether, by progressively extending the code while retaining both global and local context, the progressive skill orchestration helps preserve cross-skill dependencies and thereby improves the coherence and reliability of skill composition.

\subsection{Dataset Synthesis and SLM Training} \label{sec:distillation}

To enable SLMs to perform compositional reasoning, we use LLM teachers to generate the decomposition and orchestration datasets to distill the LLMs' strong compositional reasoning into SLMs. Additionally, since the performance of SLMs is strongly determined by the training data quality, to ensure data correctness, we go beyond direct data generation by introducing a refinement stage following~\cite{LLM-simulator} to evaluate and refine the synthesized data.

For dataset construction, we first configure an LLM to employ the \(G\) and induce a set of task instructions \(T_{train}\), then configure LLMs to synthesize the training examples of the compositional reasoning process of task decomposition and skill orchestration.

\subsubsection{Decomposition Dataset Synthesis}
The decomposition dataset is constructed to enable the SLM to decompose a task into subtasks and select appropriate skills. For each training instruction \(T^{n}_{train}\), an LLM is configured to decompose the task based on the metadata of all available skills \(\mathcal{M}(\mathcal{S})\) and generate a plan \(p^{n}\). The subtask descriptions in \(p^{n}\) are required to be concise and focused on the specific subtask operation, while the selected skill is chosen to be the most appropriate for accomplishing the subtask. Beyond task-subtask mapping, the metadata of all available skills is also included in the dataset to provide additional context for the SLMs during skill selection. The resulting decomposition dataset is:
\[
\mathcal{D}_{\mathrm{dec}}
=
\{(T^{n}_{train},\mathcal{M}(\mathcal{S}),p^{n})\}_{n=1}^{N}.
\]

\subsubsection{Orchestration Dataset Synthesis}
The orchestration dataset is constructed to enable the SLM to accomplish each subtask by orchestrating the selected skills into executable robot operation code segments. For every \(p^{{n}}\), we iterate through its steps. At step \(i\), the LLM receives the information \(x_i^{n}\) that the orchestration SLM will receive at inference time: \(T^{n}_{train}\), \(u^{n}_i\), \(b(s_{j_i})\), \(\mathcal{A}\), and the previously generated code \(C^{n}_{1:i-1}\). The LLM then generates the corresponding code segment \(c_{i}^{n}\). We then construct a dataset in which each step of the plan is treated as one supervised example:
\[
\mathcal D_{\mathrm{orch}}
=\{(x_i^{n},c_i^{n})\}_{n=1,i=1}^{N}
\]

\subsubsection{SLM Training}
We employ LoRA~\cite{lora} to reduce GPU memory usage and training time. In particular, we train two separate LoRA adapters for the decomposition and orchestration SLMs because their input/output spaces and context requirements are different. The decomposition SLM agent trained with \(\mathcal{D}_{\mathrm{dec}}\) solves task decomposition, i.e., given a task instruction, it generates \(p\). While the orchestration SLM agent trained with \(\mathcal{D}_{\mathrm{orch}}\) solves skill composition, i.e., given \(p\), it orchestrates the skills into executable robot operation code \(C\).

\section{Experiments} \label{sec:experiment}

\subsection{Experimental Setup} \label{sec:experiment setup}
\begin{figure}[htbp]
  \centering
  \begin{minipage}[t]{0.45\linewidth}
    \centering
    \includegraphics[width=\linewidth]{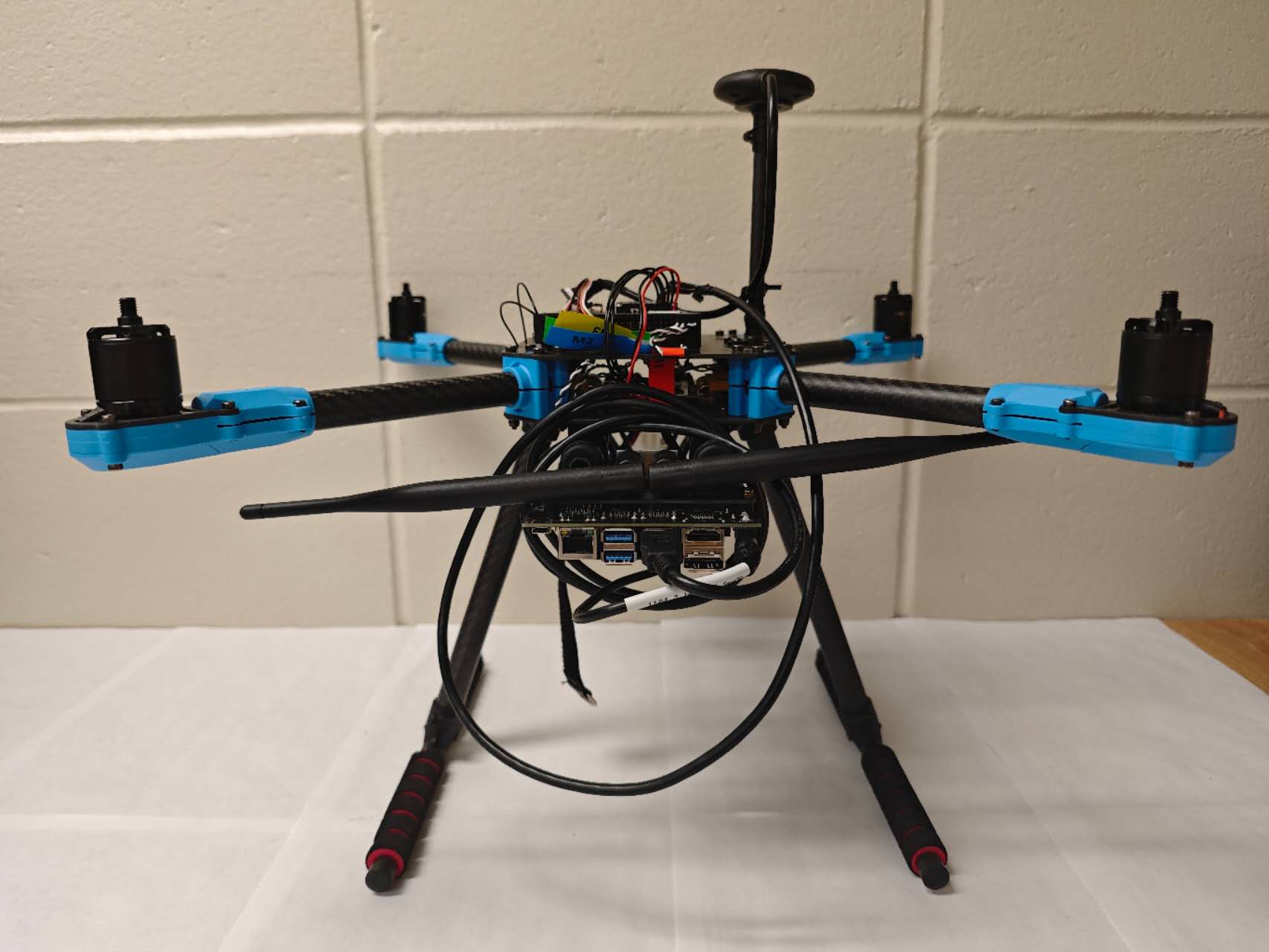}
  \end{minipage}%
  \hfill
  \begin{minipage}[t]{0.45\linewidth}
    \centering
    \includegraphics[width=\linewidth]{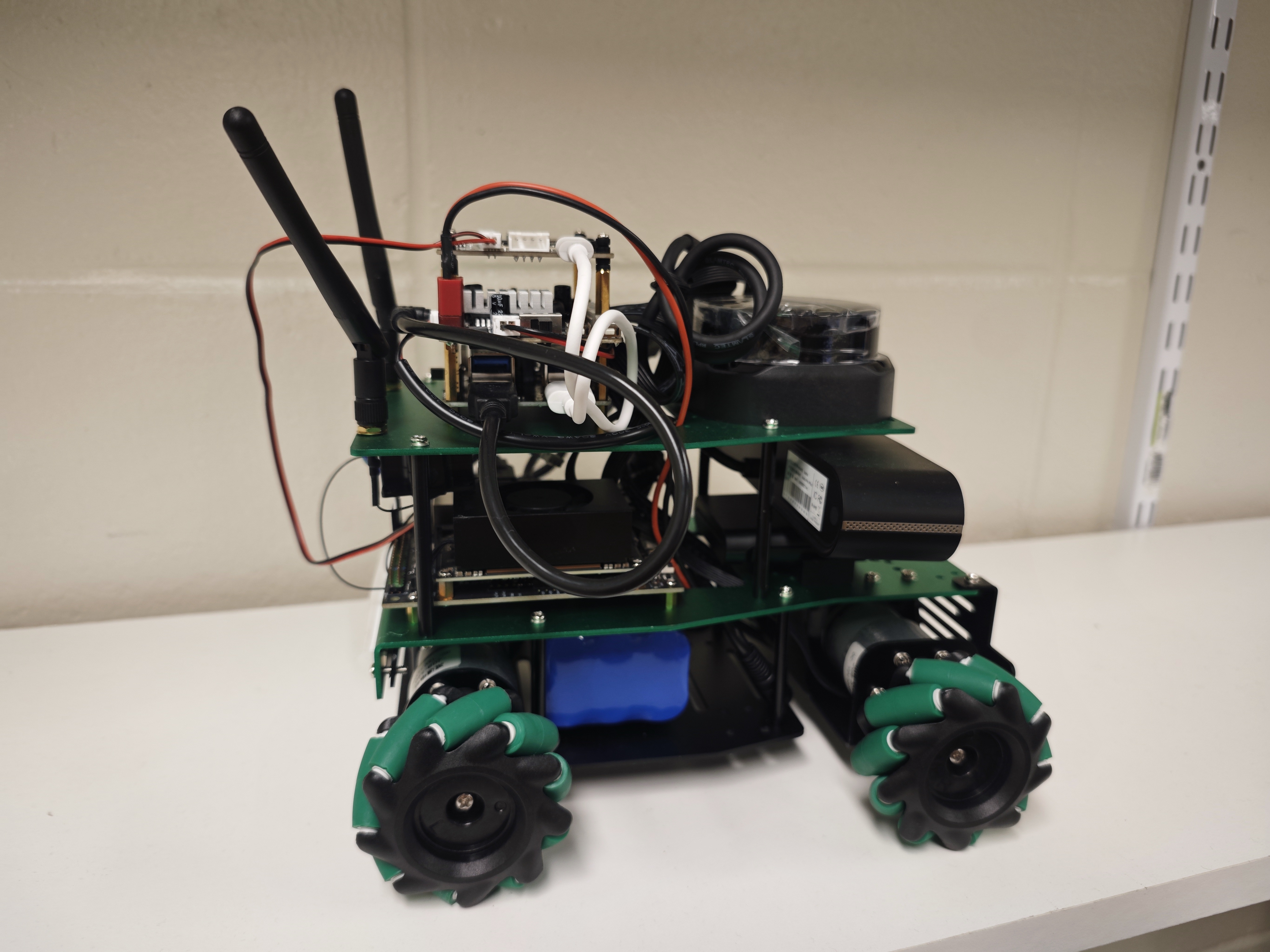}
  \end{minipage}
  \caption{UAV and ground vehicle used in the experiment.}
  \label{fig:two-images}
\end{figure}

The robots used in our experiment include a Holybro X500 V2 UAV with a Pixhawk 6X flight controller and a ROSMASTER X3 ground vehicle with ROS, as presented in Fig.~\ref{fig:two-images}. Both robots are equipped with Jetson Orin NX (16 GB) for the deployment of Skill-SLM. We evaluate four SLMs: Llama-3.2-3B-Instruct, Llama-3.1-8B-Instruct, Qwen3-4B, and Qwen3-8B~\cite{llama3, qwen3}. Model LoRA~\cite{lora} training is conducted on a server equipped with four NVIDIA H200 GPUs. For inference, we employ \(temperature=0\) and the generation is limited to 256 tokens for task decomposition and 2,048 tokens for skill orchestration. We then quantized each SLM to 4-bit ``Q4\_K\_M'' for onboard deployment on the robots. Detailed experiment settings are provided in the following sections.

\subsubsection{\textbf{Skill Library Construction}} The LLM CFG constructor is configured with OpenAI GPT-5.5~\cite{GPT-5.5}. We configure the LLM to induce a CFG from a corpus of 50 task instructions, each requiring the robot to execute a sequence of operations, as illustrated in Fig.~\ref{fig:skill-library-construction}. The resulting CFG identifies a set of skills, such as object selection, movements, and inspections. The corresponding skill library is constructed following the format described in Section~\ref{sec:skill}.

\subsubsection{\textbf{Training Set}} \label{sec:training set} We configure OpenAI GPT-5.5~\cite{GPT-5.5} as LLM teachers to construct the datasets for SLM training. To begin with, an LLM configured with the CFG from the skill library construction stage induces 300 robot task instructions. Then, an LLM teacher decomposes the task instructions into subtasks and selects the skills, resulting in 300 task-subtask-skill pairs in the decomposition dataset. Later, we configure another LLM teacher based on~\cite{LLM-simulator} to ground these subtasks in operational code, yielding 963 subtask-code pairs in the orchestration dataset.

\begin{table}[htbp]
    \centering
    \caption{Example Task Instructions in the Experiments.}
    \begin{tabular}{l|p{5.5cm}}
        \toprule
        \textbf{Task Set} & \multicolumn{1}{c}{\textbf{Example Task Instructions}} \\
        \midrule
 Seen-Pattern & ``\textcolor{gray}{\textbf{Find the nearest}} car and \textcolor{gray}{\textbf{inspect}} it.''\\
        & ``\textcolor{gray}{\textbf{Go to}} the bridge and \textcolor{gray}{\textbf{return home}}.''\\
        \midrule
 New-Combination & ``\textcolor{gray}{\textbf{Find the nearest}} bridge, \textcolor{gray}{\textbf{inspect}} it, and \textcolor{gray}{\textbf{return home}}.''\\
        \midrule
 New-Capability & ``\textcolor{gray}{\textbf{Find the nearest}} tree, \textcolor{gray}{\textbf{inspect}} it, then repeat the \textcolor{gray}{\textbf{\textit{inspection reversely}}} for opposite-perspective coverage.''\\
        \bottomrule
    \end{tabular}
    \label{tab:test tasks}
\end{table}

\subsubsection{\textbf{Test Task Sets}} \label{sec:dataset}

Following prior work on robot operational tasks~\cite{Ro-SLM}, where SLMs interpret natural language instructions and generate corresponding executable robot operation code, we construct three task sets with different settings. Each task set contains 50 tasks, with examples provided in Table~\ref{tab:test tasks}. The three task sets are defined as:

\underline{Seen-Pattern:} This set contains tasks that are well represented in the distillation dataset, where each test task has at least three in-distribution examples from which the SLM can learn. This task set is designed to evaluate whether the SLM effectively learned the knowledge and capabilities. 

\underline{New-Combination:} This set contains tasks that have limited in-distribution examples or partial coverage of the full task pattern in the distillation dataset; however, they can be solved by recombining learned knowledge. This set measures whether the SLM can generalize to more tasks through compositional reasoning.

\underline{New-Capability:} This set contains tasks that have capabilities not provided during distillation and can be solved by combining learned knowledge and adding new skills. For example, the \textit{inspect reversely} was not seen in the distillation dataset. This set evaluates whether the SLM can reliably extend its compositional reasoning to new tasks with skill library augmentation, without additional distillation.

\subsubsection{\textbf{Baselines}} \label{sec:baselines}

We compare our Skill-SLM framework with the following baselines; each baseline uses the same 300 task instructions as the training set in Section~\ref{sec:training set}.

\underline{Ro-SLM:} This baseline proposes a framework that enables an SLM to reason about the task instruction and produce code for robot operations~\cite{Ro-SLM}, where the SLM generates the complete code for a given task instruction end-to-end.

\underline{PRISM:} This baseline distills the LLM's knowledge into the SLM to make decisions and generate high-level plans for a given task instruction~\cite{PRISM}. Moreover, to evaluate PRISM in our robot execution setting, we further construct a compiler that translates the generated plans into executable code.

Additionally, to ensure a fair comparison for experiments on this task set, we provide the knowledge of new skills in the Ro-SLM~\cite{Ro-SLM} system prompt and add functions in the PRISM~\cite{PRISM} compiler.

\subsubsection{\textbf{Evaluation}} \label{sec:metrics}
Given a robot task, the ground truth is represented as a sequence of robot state transitions. Each state transition is a vector: \([x, y, z, \theta]\), where \(x\), \(y\), and \(z\) denote the robot's position changes in the North, East, and Down axes, and \(\theta\) represents yaw rotation. The corresponding robot operation is considered successful if the state transition matches the ground truth transition. To filter noise that could affect evaluation, we set a tolerance threshold of 1 meter for each positional parameter and 4 degrees for the yaw rotation in our experiments.

We evaluate performance using \textbf{Operation Success Rate (OSR)} and \textbf{Task Success Rate (TSR)}, and repeat the experiments three times:

\underline{Operation Success Rate} Given task \(T_i\), the $\mathrm{OSR}_i$ of task \(i\) is defined as $\mathrm{OSR}_i = \frac{|A_i^{\mathrm{correct}}|}{|L_i^{\mathrm{gt}}|}$, where $|L_i^{\mathrm{gt}}|$ is the length of the ground-truth transition sequence and $|A_i^{\mathrm{correct}}|$ is the number of correctly matched transitions. The OSR captures partial progress and provides insight into whether the robot operation follows the required intermediate execution steps.

\underline{Task Success Rate} Captures task-level performance to measure whether a task is fully successfully completed. Thus, a task \(T_i\) is considered \(\mathrm{TSR}_i = 1\) if and only if \(\mathrm{OSR}_i = 1\).

\begin{table*}[htbp]
    \centering
    \caption{Results on the UAV task sets.}
    \begin{tabular}{ll | c c | c c | c c}
        \toprule
        \multicolumn{2}{c}{ } & \multicolumn{2}{|c}{Seen-Pattern} & \multicolumn{2}{|c}{New-Combination} & \multicolumn{2}{|c}{New-Capability}  \\
        \midrule
        \textbf{Method} & \textbf{Model} & \textbf{TSR} & \textbf{OSR} & \textbf{TSR} & \textbf{OSR} & \textbf{TSR} & \textbf{OSR} \\
        \midrule
        \textbf{Ro-SLM} \cite{Ro-SLM}
        & Llama-3.2-3B-Instruct & $96.0\%$ & $96.4\%$ & $44.0\%$ & $70.8\%$ & $6.0\%$ & $34.1\%$ \\
        & Llama-3.1-8B-Instruct & $96.0\%$ & $98.0\%$ & $28.0\%$ & $62.3\%$ & $26.0\%$ & $52.9\%$ \\
        & Qwen3-4B & $88.0\%$ & $95.4\%$ & $44.0\%$ & $75.8\%$ & $34.0\%$ & $54.3\%$ \\
        & Qwen3-8B & $84.0\%$ & $89.6\%$ & $34.0\%$ & $69.1\%$ & $30.0\%$ & $52.1\%$ \\
        \midrule
        \textbf{PRISM} \cite{PRISM}
        & Llama-3.2-3B-Instruct & $78.0\%$ & $90.3\%$ & $28.0\%$ & $75.8\%$ & $56.0\%$ & $81.4\%$ \\
        & Llama-3.1-8B-Instruct & $80.0\%$ & $91.4\%$ & $34.0\%$ & $83.4\%$ & $66.0\%$ & $90.5\%$ \\
        & Qwen3-4B & $80.0\%$ & $91.4\%$ & $34.0\%$ & $83.4\%$ & $72.0\%$ & $93.4\%$ \\
        & Qwen3-8B & $84.0\%$ & $93.2\%$ & $32.0\%$ & $77.7\%$ & $76.0\%$ & $95.3\%$ \\
        \midrule
        \textbf{Skill-SLM}
        & Llama-3.2-3B-Instruct & $98.0\%$ & $98.0\%$ & $94.0\%$ & $96.7\%$ & $76.0\%$ & $77.7\%$ \\
        & Llama-3.1-8B-Instruct & $100.0\%$ & $100.0\%$ & $98.0\%$ & $99.3\%$ & $92.0\%$ & $92.0\%$ \\
        & Qwen3-4B & $100.0\%$ & $100.0\%$ & $98.0\%$ & $98.7\%$ & $92.0\%$ & $93.1\%$ \\
        & Qwen3-8B & $100.0\%$ & $100.0\%$ & $100.0\%$ & $100.0\%$ & $94.0\%$ & $95.2\%$ \\
        \bottomrule
    \end{tabular}
    \label{tab:overall results}
\end{table*}

\subsection{Results} \label{sec:result}

\subsubsection{\textbf{Overall Result}} \label{sec:overall result}
Table~\ref{tab:overall results} reports the results on the UAV task sets. Overall, Skill-SLM achieves the highest average TSR on all three task sets, demonstrating the effectiveness of the skill-based task-solving framework in diverse scenarios. On Seen-Pattern tasks, all methods achieve relatively strong performance because these tasks are well represented during training. Skill-SLM further achieves nearly perfect performance, with three of the four SLMs reaching $100.0\%$ TSR and OSR, indicating that reformulating robot operation as task decomposition and skill composition also improves the reliability of in-distribution task execution.

On New-Combination tasks, both PRISM and Ro-SLM exhibit substantial performance degradation, whereas Skill-SLM maintains performance close to that on Seen-Pattern tasks. This result indicates that learning complete task-solution mappings is insufficient when generalizing to tasks that require unseen combinations of learned capabilities. By decomposing the task and composing reusable skills, Skill-SLM can instead generalize the learned capabilities to new operation combinations. On New-Capability tasks, Skill-SLM also achieves the highest TSR after new skills are introduced without additional distillation, demonstrating that its compositional reasoning can be extended through skill augmentation. Moreover, PRISM's substantially higher OSR than TSR indicates that it can generate correct intermediate operations but fails to reliably compose operations into a successful task execution. In comparison, the compositional reasoning of Skill-SLM enables the SLM to apply skills and compose them into coherent executions, resulting in more reliable task success. For the few configurations where TSR equals OSR, failed tasks contain syntax errors that prevent execution, resulting in both OSR and TSR equally zero.

From a model-wise perspective, Skill-SLM consistently achieves strong performance across different SLM families and sizes. Within each model family, the larger model achieves stronger performance, demonstrating that larger models possess greater task-solving capabilities. Additionally, the Qwen models outperform their similarly sized Llama models, with Qwen3-8B achieving the strongest overall performance.

\begin{table*}[htbp]
    \centering
    \caption{Ablation results on the UAV task sets.}
    \begin{tabular}{ll | c c | c c | c c}
        \toprule
        \multicolumn{2}{c}{ } & \multicolumn{2}{|c}{Seen-Pattern} & \multicolumn{2}{|c}{New-Combination} & \multicolumn{2}{|c}{New-Capability}  \\
        \midrule
        \textbf{Ablation} & \textbf{Model} & \textbf{TSR} & \textbf{OSR} & \textbf{TSR} & \textbf{OSR} & \textbf{TSR} & \textbf{OSR} \\
        \midrule
        \textbf{Direct Skill} & Llama-3.2-3B-Instruct & $92.0\%$ & $93.5\%$ & $90.0\%$ & $93.0\%$ & $46.0\%$ & $57.1\%$ \\
        \textbf{Extraction} & Llama-3.1-8B-Instruct & $96.0\%$ & $97.2\%$ & $92.0\%$ & $94.7\%$ & $68.0\%$ & $75.0\%$ \\
        & Qwen3-4B & $94.0\%$ & $96.0\%$ & $98.0\%$ & $99.5\%$ & $58.0\%$ & $67.7\%$ \\
        & Qwen3-8B & $98.0\%$ & $99.0\%$ & $98.0\%$ & $99.0\%$ & $88.0\%$ & $91.0\%$ \\
        
        \midrule
        \textbf{No} & Llama-3.2-3B-Instruct & $4.0\%$ & $11.6\%$ & $0.0\%$ & $3.0\%$ & $0.0\%$ & $5.1\%$ \\
        \textbf{Distillation} & Llama-3.1-8B-Instruct & $32.0\%$ & $39.5\%$ & $0.0\%$ & $8.8\%$ & $2.0\%$ & $10.0\%$ \\
        & Qwen3-4B & $46.0\%$ & $50.9\%$ & $32.0\%$ & $49.4\%$ & $26.0\%$ & $39.1\%$ \\
        & Qwen3-8B & $70.0\%$ & $72.2\%$ & $50.0\%$ & $64.2\%$ & $58.0\%$ & $67.3\%$ \\

        \midrule
        \textbf{One-Shot} & Llama-3.2-3B-Instruct & $90.0\%$ & $90.0\%$ & $66.0\%$ & $80.7\%$ & $72.0\%$ & $83.5\%$ \\
        \textbf{Orchestration} & Llama-3.1-8B-Instruct & $98.0\%$ & $99.0\%$ & $84.0\%$ & $89.6\%$ & $92.0\%$ & $94.1\%$ \\
        & Qwen3-4B & $98.0\%$ & $99.0\%$ & $94.0\%$ & $98.0\%$ & $90.0\%$ & $92.5\%$ \\
        & Qwen3-8B & $96.0\%$ & $97.2\%$ & $94.0\%$ & $97.0\%$ & $92.0\%$ & $98.1\%$ \\
        \bottomrule
    \end{tabular}
    \label{tab:ablation results}
\end{table*}

\begin{table*}[t]
    \centering
    \caption{Skill-SLM results on the ground vehicle task sets.}
    \begin{tabular}{l | c c | c c | c c}
        \toprule
        \multicolumn{1}{c}{ }& \multicolumn{2}{|c|}{Seen-Pattern} & \multicolumn{2}{c}{New-Combination} & \multicolumn{2}{|c}{New-Capability}  \\
        \midrule
        \textbf{Model} & \textbf{TSR} & \textbf{OSR} & \textbf{TSR} & \textbf{OSR} & \textbf{TSR} & \textbf{OSR} \\
        \midrule
 Llama-3.2-3B-Instruct & $90.0\%$ & $94.0\%$ & $85.0\%$ & $88.3\%$ & $70.0\%$ & $77.3\%$ \\
 Llama-3.1-8B-Instruct & $100.0\%$ & $100.0\%$ & $90.0\%$ & $92.0\%$ & $85.0\%$ & $91.1\%$ \\
 Qwen3-4B & $100.0\%$ & $100.0\%$ & $95.0\%$ & $97.0\%$ & $80.0\%$ & $97.2\%$ \\
 Qwen3-8B & $100.0\%$ & $100.0\%$ & $95.0\%$ & $97.0\%$ & $100.0\%$ & $100.0\%$ \\
        \bottomrule
    \end{tabular}
    \label{tab:ground vehicle results}
\end{table*}

\subsubsection{\textbf{Ablation Study}} \label{sec:ablation result}

The ablation study evaluates three variants: \underline{Direct Skill Extraction}, the LLM directly constructs a CFG that extracts skills from the given task corpus without operational skill awareness; \underline{No Distillation}, it retains the Skill-SLM inference workflow but uses the base SLMs without fine-tuning; and \underline{One-Shot Orchestration}, it replaces progressive orchestration with a single call that generates the complete operation code from plan \(p\). The results are presented in Table~\ref{tab:ablation results}.

\underline{Direct Skill Extraction} Direct skill extraction significantly falls behind on New-Capability tasks. The degradation indicates that directly extracting skills from the task corpus does not reliably identify consistent and complete operational skills as the skill library expands. Redundant skills make it difficult for the SLM to select and compose newly introduced capabilities correctly. This result indicates that directly extracted skills cannot achieve reliable performance as Skill-SLM over various task scenarios. Our operational skill awareness CFG is therefore important for constructing a skill library that supports reliable skill-driven reasoning and execution across diverse scenarios.

\underline{No Distillation} Removing distillation causes the largest degradation. Although Qwen3-8B retains moderate performance, it remains substantially less reliable than the distilled Skill-SLM. These results indicate that system prompts and skill documents alone are insufficient for base SLMs to perform skill-driven reasoning reliably. Therefore, learning task decomposition and skill orchestration from teacher-generated examples can transfer the compositional reasoning ability from the LLM and enable the base SLM to apply it effectively during robot execution.

\underline{One-Shot Orchestration} The one-shot approach processes the complete plan and is overwhelmed by the task context and skill documents, making it difficult to maintain coherence for successful task execution. The TSR degradation indicates that one-shot orchestration often produces correct intermediate operations but fails to maintain coherence for successful task execution. Our progressive orchestration instead generates code subtask-by-subtask while preserving prior context, improving the coherence of the generated code and therefore enhancing the success of task execution.

\subsection{Ground Vehicle} \label{sec:ground vehicle result}
To demonstrate the transferability of Skill-SLM, we further evaluate it on ground vehicle operations. We follow the same experimental setup as in the UAV experiments to build the training dataset based on the ground vehicle operations. We tested on three ground vehicle task sets analogous to the UAV in Section~\ref{sec:dataset}, with 20 tasks each.

Table~\ref{tab:ground vehicle results} reports the results on the ground vehicle task sets. On Seen-Pattern tasks, all models achieve strong performance, demonstrating that Skill-SLM can reliably learn and execute ground vehicle operations. Skill-SLM also maintains consistent performance on New-Combination tasks, indicating that the SLMs can effectively reuse and compose learned ground vehicle skills to solve novel task combinations. On New-Capability tasks, although the tasks require integrating newly introduced skills, Skill-SLM still exhibits robust performance, highlighting its ability to effectively adapt and incorporate new capabilities without additional training. From a model-wise perspective, Llama-3.2-3B-Instruct exhibits slightly lower performance due to its smaller scale, whereas Qwen3-8B achieves the strongest overall performance. These results are consistent with the UAV experiments and demonstrate that Skill-SLM can generalize its skill-based compositional reasoning across different robot platforms and scenarios.

\subsection{Latency and Memory Usage} \label{sec:resource usage}

Skill-SLM is designed for language model driven task planning and robot operation rather than real-time low-level control. Accordingly, we evaluate its onboard inference efficiency on an embedded robotic platform. All models are deployed on an NVIDIA Jetson Orin NX with 16~GB of memory using 4-bit Q4\_K\_M quantization and Ollama. The results reveal a clear trade-off between model scale and deployment efficiency. Llama-3.2-3B-Instruct achieves the highest output throughput at 18.3 tokens/s and requires only 4.3~GB of peak memory, while Qwen3-4B maintains a high throughput of 15.5 tokens/s with 6.0~GB of memory usage. The two 8B models exhibit similar throughput, with Llama-3.1-8B-Instruct and Qwen3-8B generating 10.6 and 10.7 tokens/s while requiring 9.5 and 10.4~GB of peak memory, respectively. Additionally, when considered together with the task success results, Qwen3-4B provides a favorable balance between task-solving performance and deployment efficiency, achieving performance comparable to the 8B models while requiring less memory and providing higher throughput. Meanwhile, for deployment on memory-constrained robots, Llama-3.2-3B-Instruct offers the most efficient configuration.

\section{Conclusion}

In this paper, we introduced Skill-SLM, an agent skill-driven framework for reliable onboard robot operation with SLMs. Skill-SLM reformulates robot operation as a task-decomposition and skill-composition problem. It constructs the skill library with an operational skill aware CFG, distills LLM knowledge and capabilities into SLM agents, and progressively orchestrates selected skills into coherent executable robot operations. Experiments on UAV tasks demonstrate that Skill-SLM substantially outperforms the baselines, particularly on tasks that require recombining learned capabilities. In addition, the ground vehicle results demonstrate the generality of Skill-SLM across different robot platforms. Overall, Skill-SLM provides a promising approach to reliable and generalizable onboard task planning and robot operations. 

% \addtolength{\textheight}{-12cm} 

%%%%%%%%%%%%%%%%%%%%%%%%%%%%%%%%%%%%%%%%%%%%%%%%%%%%%%%%%%%%%%%%%%%%%%%%%%%%%%%%

%%%%%%%%%%%%%%%%%%%%%%%%%%%%%%%%%%%%%%%%%%%%%%%%%%%%%%%%%%%%%%%%%%%%%%%%%%%%%%%%



% \section*{ACKNOWLEDGMENT}

\bibliographystyle{IEEEtran}
\bibliography{ref}

\end{document}